\documentclass[11pt,a4paper]{article}
\usepackage[hyperref]{acl2017}
\usepackage{times}
\usepackage{latexsym}
\usepackage{helvet}
\usepackage{courier}
\usepackage{multirow}
\usepackage{graphicx}
\usepackage{amsfonts}
\usepackage{stmaryrd}
\usepackage{bm}
\usepackage{amsfonts}
\usepackage{algorithm}
\usepackage{algpseudocode}
\usepackage{color}
\usepackage{bbm}
\usepackage{amsmath}

\usepackage{url}

\newtheorem{definition}{Assumption}

\def\argmax{\mathop{\rm argmax}}
\def\argmin{\mathop{\rm argmin}}

\aclfinalcopy 
\def\aclpaperid{779} 

\title{A Teacher-Student Framework for \\
Zero-Resource Neural Machine Translation}
\author{{Yun Chen$^\dagger$, Yang Liu$^\ddagger$, Yong Cheng$^+$, Victor O.K. Li$^\dagger$}\\
	$^\dagger$Department of Electrical and Electronic Engineering, The University of Hong Kong\\ $^\ddagger$State Key Laboratory of Intelligent Technology and Systems \\
	Tsinghua National Laboratory for Information Science and Technology \\
	Department of Computer Science and Technology, Tsinghua University, Beijing, China\\
	Jiangsu Collaborative Innovation Center for Language Competence, Jiangsu, China \\
	$^+$Institute for Interdisciplinary Information Sciences, Tsinghua University, Beijing, China\\
\tt yun.chencreek@gmail.com; liuyang2011@tsinghua.edu.cn;\\ \tt  chengyong3001@gmail.com; vli@eee.hku.hk}

\begin{document}
\maketitle

\begin{abstract}
While end-to-end neural machine translation (NMT) has made remarkable progress recently, it still suffers from the data scarcity problem for low-resource language pairs and domains. In this paper, we propose a method for zero-resource NMT by assuming that parallel sentences have close probabilities of generating a sentence in a third language. Based on this assumption, our method is able to train a source-to-target NMT model (``student'') without parallel corpora available, guided by an existing pivot-to-target NMT model (``teacher'') on a source-pivot parallel corpus. Experimental results show that the proposed method significantly improves over a baseline pivot-based model by +3.0 BLEU points across various language pairs.
\end{abstract}

\section{Introduction}
\begin{figure*}[!t]
	\centering
	\includegraphics[width=6.3in]{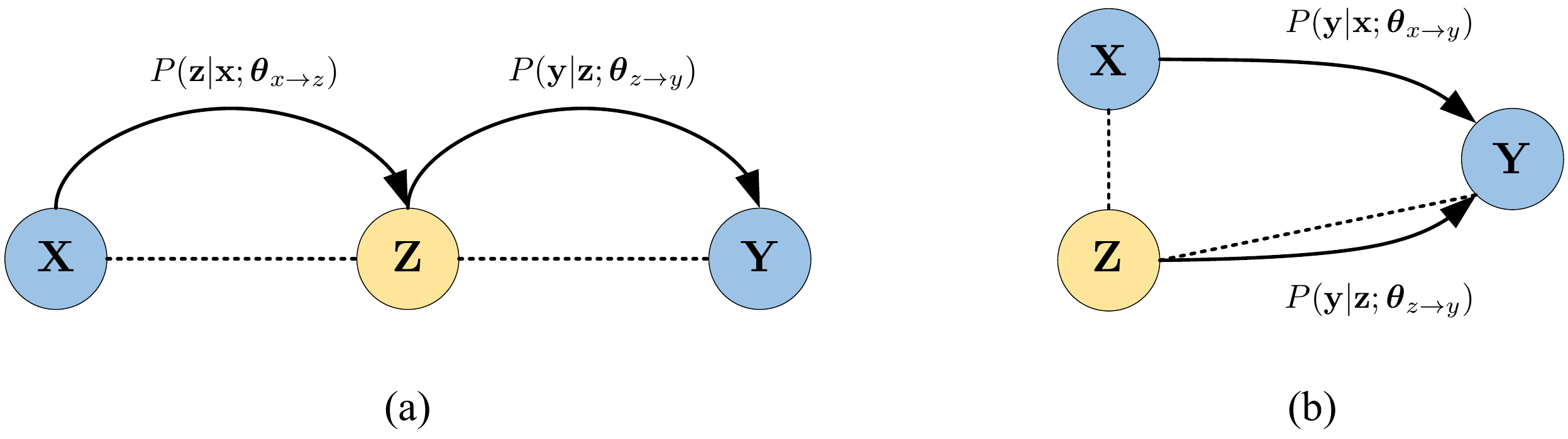}
	\caption{(a) The pivot-based approach and (b) the teacher-student approach to zero-resource neural machine translation. $\mathbf{X}$, $\mathbf{Y}$, and $\mathbf{Z}$ denote source, target, and pivot languages, respectively. We use a dashed line to denote that there is a parallel corpus available for the connected language pair. Solid lines with arrows represent translation directions. The pivot-based approach leverages a pivot to achieve indirect source-to-target translation: it first translates $\mathbf{x}$ into $\mathbf{z}$, which is then translated into $\mathbf{y}$. Our training algorithm is based on the translation equivalence assumption: if $\mathbf{x}$ is a translation of $\mathbf{z}$, then $P(\mathbf{y}|\mathbf{x}; \bm{\theta}_{x \rightarrow y})$ should be close to $P(\mathbf{y}|\mathbf{z}; \bm{\theta}_{z \rightarrow y})$. Our approach directly trains the intended source-to-target model $P(\mathbf{y}|\mathbf{x}; \bm{\theta}_{x \rightarrow y})$ (``student'') on a source-pivot parallel corpus, with the guidance of an existing pivot-to-target model $P(\mathbf{y}|\mathbf{z}; \hat{\bm{\theta}}_{z \rightarrow y})$ (``teacher'').   } \label{fig:1}
\end{figure*}
Neural machine translation (NMT) \cite{Kalchbrenner2013RecurrentCT,Sutskever2014SequenceTS,Bahdanau2014NeuralMT}, which directly models the translation process in an end-to-end way, has attracted intensive attention from the community. Although NMT has achieved state-of-the-art translation performance on resource-rich language pairs such as English-French and German-English \cite{Luong2015AddressingTR,Jean2015OnUV,Wu2016GooglesNM,Johnson2016GooglesMN}, it still suffers from the unavailability of large-scale parallel corpora for translating low-resource languages. Due to the large parameter space, neural models usually learn poorly from low-count events, resulting in a poor choice for low-resource language pairs. Zoph et al. \shortcite{Zoph2016TransferLF} indicate that NMT obtains much worse translation quality than a statistical machine translation (SMT) system on low-resource languages.

As a result, a number of authors have endeavored to explore methods for translating language pairs without parallel corpora available. These methods can be roughly divided into two broad categories: {\em multilingual} and {\em pivot-based}. Firat et al. \shortcite{Firat2016ZeroResourceTW} present a multi-way, multilingual model with shared attention to achieve zero-resource translation. They fine-tune the attention part using pseudo bilingual sentences for the zero-resource language pair. Another direction is to develop a universal NMT model in multilingual scenarios \cite{Johnson2016GooglesMN,Ha2016TowardMN}. They use parallel corpora of multiple languages to train one single model, which is then able to translate a language pair without parallel corpora available. Although these approaches prove to be effective, the combination of multiple languages in modeling and training leads to increased complexity compared with standard NMT.

Another direction is to achieve source-to-target NMT without parallel data via a \textit{pivot}, which is either text \cite{Cheng2016NeuralMT} or image \cite{Nakayama2016ZeroresourceMT}. Cheng et al. \shortcite{Cheng2016NeuralMT} propose a pivot-based method for zero-resource NMT: it first translates the source language to a pivot language, which is then translated to the target language. Nakayama and Nishida \shortcite{Nakayama2016ZeroresourceMT} show that using multimedia information as pivot also benefits zero-resource translation. However, pivot-based approaches usually need to divide the decoding process into two steps, which is not only more computationally expensive, but also potentially suffers from the error propagation problem \cite{Zhu2013ImprovingPS}.

In this paper, we propose a new method for zero-resource neural machine translation. Our method assumes that parallel sentences should have close probabilities of generating a sentence in a third language. To train a source-to-target NMT model without parallel corpora available (``student''), we leverage an existing pivot-to-target NMT model (``teacher'') to guide the learning process of the student model on a source-pivot parallel corpus. Compared with pivot-based approaches \cite{Cheng2016NeuralMT}, our method allows direct parameter estimation of the intended NMT model, without the need to divide decoding into two steps. This strategy not only improves efficiency but also avoids error propagation in decoding. Experiments on the Europarl and WMT datasets show that our approach achieves significant improvements in terms of both translation quality and decoding efficiency over a baseline pivot-based approach to zero-resource NMT on Spanish-French and German-French translation tasks.

\section{Background}

Neural machine translation \cite{Sutskever2014SequenceTS,Bahdanau2014NeuralMT} advocates the use of neural networks to model the translation process in an end-to-end manner. As a data-driven approach, NMT treats parallel corpora as the major source for acquiring translation knowledge.

Let $\mathbf{x}$ be a source-language sentence and $\mathbf{y}$ be a target-language sentence. We use $P(\mathbf{y}|\mathbf{x}; \bm{\theta}_{x\rightarrow y})$ to denote a source-to-target neural translation model, where $\bm{\theta}_{x \rightarrow y}$ is a set of model parameters. Given a source-target parallel corpus $D_{x,y}$, which is a set of parallel source-target sentences, the model parameters can be learned by maximizing the log-likelihood of the parallel corpus:
\begin{eqnarray}
\hat{\bm{\theta}}_{x \rightarrow y} = \argmax_{\bm{\theta}_{x \rightarrow y}} \Bigg\{  \sum_{\langle \mathbf{x}, \mathbf{y} \rangle \in D_{x,y} } \log P(\mathbf{y}|\mathbf{x}; \bm{\theta}_{x \rightarrow y}) \Bigg\}. \nonumber
\end{eqnarray}

Given learned model parameters $\hat{\bm{\theta}}_{x \rightarrow y}$, the decision rule for finding the translation with the highest probability for a source sentence $\mathbf{x}$ is given by
\begin{eqnarray}
\hat{\mathbf{y}} = \argmax_{\mathbf{y}}\Bigg\{ P(\mathbf{y}|\mathbf{x}; \hat{\bm{\theta}}_{x \rightarrow y}) \Bigg\}. \label{std:decode}
\end{eqnarray}

As a data-driven approach, NMT heavily relies on the availability of large-scale parallel corpora to deliver state-of-the-art translation performance \cite{Wu2016GooglesNM,Johnson2016GooglesMN}. Zoph et al. \shortcite{Zoph2016TransferLF} report that NMT obtains much lower BLEU scores than SMT if only small-scale parallel corpora are available. Therefore, the heavy dependence on the quantity of training data poses a severe challenge for NMT to translate zero-resource language pairs.

Simple and easy-to-implement, pivot-based methods have been widely used in SMT for translating zero-resource language pairs \cite{de2006catalan, Cohn2007MachineTB, Utiyama2007ACO, Wu2007PivotLA, Bertoldi2008PhrasebasedSM, Wu2009RevisitingPL, Zahabi2013UsingCV, Kholy2013LanguageIC}. As pivot-based methods are agnostic to model structures, they have been adapted to NMT recently \cite{Cheng2016NeuralMT,Johnson2016GooglesMN}.

Figure \ref{fig:1}(a) illustrates the basic idea of pivot-based approaches to zero-resource NMT \cite{Cheng2016NeuralMT}. Let $\mathbf{X}$, $\mathbf{Y}$, and $\mathbf{Z}$ denote source, target, and pivot languages. We use dashed lines to denote language pairs with parallel corpora available and solid lines with arrows to denote translation directions.

Intuitively, the source-to-target translation can be indirectly modeled by bridging two NMT models via a pivot:
\begin{eqnarray}
&& P(\mathbf{y}|\mathbf{x}; \bm{\theta}_{x \rightarrow z}, \bm{\theta}_{z \rightarrow y}) \nonumber \\
& =& \sum_{\mathbf{z}}P(\mathbf{z}|\mathbf{x}; \bm{\theta}_{x \rightarrow z})P(\mathbf{y}|\mathbf{z}; \bm{\theta}_{z \rightarrow y}).
\end{eqnarray}

As shown in Figure \ref{fig:1}(a), pivot-based approaches assume that the source-pivot parallel corpus $D_{x,z}$ and the pivot-target parallel corpus $D_{z,y}$ are available.  As it is impractical to enumerate all possible pivot sentences, the two NMT models are trained separately in practice:
\begin{eqnarray}
\hat{\bm{\theta}}_{x \rightarrow z} = \argmax_{\mathbf{\theta}_{x \rightarrow z}} \Bigg\{ \sum_{\langle \mathbf{x}, \mathbf{z} \rangle \in D_{x,z} } \log P(\mathbf{z}|\mathbf{x}; \bm{\theta}_{x \rightarrow z}) \Bigg\}, \nonumber \\
\hat{\bm{\theta}}_{z \rightarrow y} = \argmax_{\mathbf{\theta}_{z \rightarrow y}} \Bigg\{ \sum_{\langle \mathbf{z}, \mathbf{y} \rangle \in D_{z,y} } \log P(\mathbf{y}|\mathbf{z}; \bm{\theta}_{z \rightarrow y}) \Bigg\}. \nonumber
\end{eqnarray}

Due to the exponential search space of pivot sentences, the decoding process of translating an unseen source sentence $\mathbf{x}$ has to be divided into two steps:
\begin{eqnarray}
\hat{\mathbf{z}} = \argmax_{\mathbf{z}}\Big\{ P(\mathbf{z}|\mathbf{x}; \hat{\bm{\theta}}_{x \rightarrow z}) \Big\}, \\
\hat{\mathbf{y}} = \argmax_{\mathbf{y}}\Big\{ P(\mathbf{y}|\hat{\mathbf{z}}; \hat{\bm{\theta}}_{z \rightarrow y}) \Big\}.
\end{eqnarray}
The above two-step decoding process potentially suffers from the error propagation problem \cite{Zhu2013ImprovingPS}: the translation errors made in the first step (i.e., source-to-pivot translation) will affect the second step (i.e., pivot-to-target translation).

Therefore, it is necessary to explore methods to directly model source-to-target translation without parallel corpora available.

\section{Approach}

\subsection{Assumptions}\label{C1}
In this work, we propose to directly model the intended source-to-target neural translation based on a teacher-student framework. The basic idea is to use a pre-trained pivot-to-target model (``teacher'') to guide the learning process of a source-to-target model (``student'') without training data available on a source-pivot parallel corpus. One advantage of our approach is that Equation (\ref{std:decode}) can be used as the decision rule for decoding, which avoids the error propagation problem faced by two-step decoding in pivot-based approaches.

As shown in Figure \ref{fig:1}(b), we still assume that a source-pivot parallel corpus $D_{x,z}$ and a pivot-target parallel corpus $D_{z,y}$ are available. Unlike pivot-based approaches, we first use the pivot-target parallel corpus $D_{z,y}$ to obtain a {\bf teacher model} $P(\mathbf{y}|\mathbf{z}; \hat{\bm{\theta}}_{z \rightarrow y})$, where $\hat{\bm{\theta}}_{z \rightarrow y}$ is a set of learned model parameters. Then, the teacher model ``teaches'' the {\bf student model} $P(\mathbf{y}|\mathbf{x}; \bm{\theta}_{x \rightarrow y})$ on the source-pivot parallel corpus $D_{x,z}$ based on the following assumptions.

\begin{definition}
If a source sentence $\mathbf{x}$ is a translation of a pivot sentence $\mathbf{z}$, then the probability of generating a target sentence $\mathbf{y}$ from $\mathbf{x}$  should be close to that from its counterpart $\mathbf{z}$. \label{a1}
\end{definition}

We can further introduce a word-level assumption:

\begin{definition}
If a source sentence $\mathbf{x}$ is a translation of a pivot sentence $\mathbf{z}$, then the probability of generating a target word $y$ from $\mathbf{x}$  should be close to that from its counterpart $\mathbf{z}$, given the already obtained partial translation $\mathbf{y}_{<j}$. \label{a2}
\end{definition}

The two assumptions are empirically verified in our experiments (see Table \ref{table:VF}). In the following subsections, we will introduce two approaches to zero-resource neural machine translation based on the two assumptions.

\subsection{Sentence-Level Teaching}\label{3.2}

Given a source-pivot parallel corpus $D_{x,z}$, our training objective based on Assumption \ref{a1} is defined as follows:
\begin{eqnarray}
\mathcal{J}_{\mathrm{SENT}}(\bm{\theta}_{x \rightarrow y}) \quad \quad \quad \quad  \quad \quad \quad \quad \quad \quad \quad \quad  \nonumber \\
= \sum_{\langle \mathbf{x}, \mathbf{z}\rangle \in D_{x,z} } \mathrm{KL}\Big(P(\mathbf{y}|\mathbf{z}; \hat{\bm{\theta}}_{z \rightarrow y}) \Big|\Big| P(\mathbf{y}|\mathbf{x}; \bm{\theta}_{x \rightarrow y})\Big), \label{eq:sent}
\end{eqnarray}
where the KL divergence sums over all possible target sentences:
\begin{eqnarray}
&& \mathrm{KL}\Big(P(\mathbf{y}|\mathbf{z}; \hat{\bm{\theta}}_{z \rightarrow y}) \Big|\Big| P(\mathbf{y}|\mathbf{x}; \bm{\theta}_{x \rightarrow y})\Big) \nonumber \\
&=& \sum_{\mathbf{y}} P(\mathbf{y}|\mathbf{z}; \hat{\bm{\theta}}_{z \rightarrow y}) \log \frac{P(\mathbf{y}|\mathbf{z}; \hat{\bm{\theta}}_{z \rightarrow y})}{P(\mathbf{y}|\mathbf{x}; \bm{\theta}_{x \rightarrow y})}.
\end{eqnarray}
As the teacher model parameters are fixed, the training objective can be equivalently written as
\begin{eqnarray}
\mathcal{J}_{\mathrm{SENT}}(\bm{\theta}_{x \rightarrow y}) \quad \quad \quad \quad \quad \quad \quad \quad \quad \quad \quad \ \ \nonumber \\
=-\sum_{\langle \mathbf{x}, \mathbf{z}\rangle \in D_{x,z} } \mathbb{E}_{ \mathbf{y}|\mathbf{z}; \hat{\bm{\theta}}_{z \rightarrow y} } \Big[ \log P(\mathbf{y}|\mathbf{x}; \bm{\theta}_{x \rightarrow y}) \Big]. \label{sentence:final}
\end{eqnarray}

In training, our goal is to find a set of source-to-target model parameters that minimizes the training objective:
\begin{eqnarray}
\hat{\bm{\theta}}_{x \rightarrow y} = \argmin_{\bm{\theta}_{x \rightarrow y}} \Big\{ \mathcal{J}_{\mathrm{SENT}}(\bm{\theta}_{x \rightarrow y}) \Big\}.
\end{eqnarray}

With learned source-to-target model parameters $\hat{\bm{\theta}}_{x \rightarrow y}$, we use the standard decision rule as shown in Equation (\ref{std:decode}) to find the translation $\hat{\mathbf{y}}$ for a source sentence $\mathbf{x}$.

However, a major difficulty faced by our approach is the intractability in calculating the gradients because of the exponential search space of target sentences. To address this problem, it is possible to construct a sub-space by either sampling \cite{Shen2016MinimumRT}, generating a $k$-best list \cite{Cheng2016SemiSupervisedLF} or mode approximation \cite{Kim2016SequenceLevelKD}. Then, standard stochastic gradient descent algorithms can be used to optimize model parameters.

\subsection{Word-Level Teaching}
Instead of minimizing the KL divergence between the teacher and student models at the sentence level, we further define a training objective  at the word level based on Assumption \ref{a2}:

\begin{eqnarray}
\mathcal{J}_{\mathrm{WORD}}(\bm{\theta}_{x \rightarrow y}) \quad \quad \quad \quad \quad \quad \quad \quad \quad \quad \quad \ \ \ \nonumber \\
= \sum_{\langle \mathbf{x}, \mathbf{z} \rangle \in D_{x,z}}\mathbb{E}_{\mathbf{y}|\mathbf{z}; \hat{\bm{\theta}}_{z\rightarrow y}}\Big[ J(\mathbf{x}, \mathbf{y}, \mathbf{z}, \hat{\bm{\theta}}_{z \rightarrow y}, \bm{\theta}_{x \rightarrow y}) \Big], \label{eq:word}
\end{eqnarray} 
where
\begin{eqnarray}
&&J(\mathbf{x}, \mathbf{y}, \mathbf{z}, \hat{\bm{\theta}}_{z \rightarrow y}, \bm{\theta}_{x \rightarrow y})  \nonumber \\
&=& \sum_{j=1}^{|\mathbf{y}|}\mathrm{KL}\Big(P(y|\mathbf{z},\mathbf{y}_{<j}; \hat{\bm{\theta}}_{z\rightarrow y}) \Big|\Big| \nonumber \\
 && \quad \quad \quad \ \ P(y|\mathbf{x}, \mathbf{y}_{<j}; \bm{\theta}_{x \rightarrow y})\Big).
\end{eqnarray}

Equation (\ref{eq:word}) suggests that the teacher model $P(y|\mathbf{z}, \mathbf{y}_{<j}; \hat{\bm{\theta}}_{z \rightarrow y})$ ``teaches'' the student model $P(y|\mathbf{x}, \mathbf{y}_{<j}; \bm{\theta}_{x \rightarrow y})$ in a word-by-word way. Note that the KL-divergence between the two models is defined at the word level:
\begin{eqnarray}
\mathrm{KL} \Big(P(y|\mathbf{z},\mathbf{y}_{<j}; \hat{\bm{\theta}}_{z\rightarrow y}) \Big|\Big| P(y|\mathbf{x}, \mathbf{y}_{<j}; \bm{\theta}_{x \rightarrow y})\Big) \nonumber \\
= \sum_{y \in \mathcal{V}_y} P(y|\mathbf{z},\mathbf{y}_{<j}; \hat{\bm{\theta}}_{z\rightarrow y}) \log \frac{P(y|\mathbf{z},\mathbf{y}_{<j}; \hat{\bm{\theta}}_{z\rightarrow y})}{P(y|\mathbf{x}, \mathbf{y}_{<j}; \bm{\theta}_{x \rightarrow y})}, \nonumber
\end{eqnarray}
where $\mathcal{V}_y$ is the target vocabulary. As the parameters of the teacher model  are fixed, the training objective can be equivalently written as:
\begin{eqnarray}
\mathcal{J}_{\mathrm{WORD}}(\bm{\theta}_{x \rightarrow y}) \quad \quad \quad \quad \quad \quad \quad \quad \quad \quad \quad \ \ \ \nonumber \\
= -\sum_{\langle \mathbf{x}, \mathbf{z} \rangle \in D_{x,z}}\mathbb{E}_{\mathbf{y}|\mathbf{z}; \hat{\bm{\theta}}_{z\rightarrow y}}\Big[ S(\mathbf{x}, \mathbf{y}, \mathbf{z}, \hat{\bm{\theta}}_{z \rightarrow y}, \bm{\theta}_{x \rightarrow y}) \Big],
\end{eqnarray} 
where
\begin{eqnarray}
S(\mathbf{x}, \mathbf{y}, \mathbf{z}, \hat{\bm{\theta}}_{z \rightarrow y}, \bm{\theta}_{x \rightarrow y}) \quad \quad \quad \ \ \ \ \nonumber \\
= \sum_{j=1}^{|\mathbf{y}|} \sum_{y \in \mathcal{V}_y}
P(y|\mathbf{z},\mathbf{y}_{<j};\hat{\bm{\theta}}_{z\rightarrow y})\times \quad \nonumber \\
 \log P(y|\mathbf{x}, \mathbf{y}_{<j}; \bm{\theta}_{x \rightarrow y}).\
\label{word:final}
\end{eqnarray}

Therefore, our goal is to find a set of source-to-target model parameters that minimizes the training objective:
\begin{eqnarray}
\hat{\bm{\theta}}_{x \rightarrow y} = \argmin_{\bm{\theta}_{x \rightarrow y}} \Big\{ \mathcal{J}_{\mathrm{WORD}}(\bm{\theta}_{x \rightarrow y}) \Big\}.
\end{eqnarray}

We use similar approaches as described in Section \ref{3.2} for approximating the full search space with sentence-level teaching. After obtaining $\hat{\bm{\theta}}_{x \rightarrow y}$, the same decision rule as shown in Equation (\ref{std:decode}) can be utilized to find the most probable target sentence $\hat{\mathbf{y}}$ for a source sentence $\mathbf{x}$.

\section{Experiments}

\subsection{Setup}\label{c2}
\begin{table}[!t]
	\centering
	\begin{tabular}{l | l | r | r | r }
		Corpus & Direction & Train & Dev. & Test  \\
		\hline \hline
		\multirow{3}{*}{Europarl} & Es$\rightarrow$ En  & 850K  & 2,000 & 2,000  \\
		& De$\rightarrow$ En  & 840K  & 2,000 & 2,000 \\
		& En$\rightarrow$ Fr  & 900K  & 2,000 & 2,000 \\
		\hline \hline
		\multirow{2}{*}{WMT} & Es$\rightarrow$ En  & 6.78M  & 3,003 & 3,003 \\
		& En$\rightarrow$ Fr  & 9.29M  & 3,003 & 3,003 \\
		
	\end{tabular}
	\caption{Data statistics. For the Europarl corpus, we evaluate our approach on Spanish-French (Es-Fr) and German-French (De-Fr) translation tasks. For the WMT corpus, we evaluate our approach on the Spanish-French (Es-Fr) translation task. English is used as a pivot language in all experiments.} \vspace{-10pt} \label{table:D1}
\end{table}
We evaluate our approach on the Europarl \cite{Koehn2005EuroparlAP} and WMT corpora. To compare with pivot-based methods, we use the same dataset as \cite{Cheng2016NeuralMT}. All the sentences are tokenized by the \verb|tokenize.perl| script. All the experiments treat English as the pivot language and French as the target language.

For the Europarl corpus, we evaluate our proposed methods on Spanish-French (Es-Fr) and German-French (De-Fr) translation tasks in a zero-resource scenario. To avoid the trilingual corpus constituted by the source-pivot and pivot-target corpora, we split the overlapping pivot sentences of the original source-pivot and pivot-target corpora into two equal parts and merge them separately with the non-overlapping parts for each language pair. The development and test sets are from WMT 2006 shared task.\footnote{http://www.statmt.org/wmt07/shared-task.html} The evaluation metric is case-insensitive BLEU \cite{Papineni2002BleuAM} as calculated by the \verb|multi-bleu.perl| script. To deal with out-of-vocabulary words, we adopt byte pair encoding (BPE) \cite{Sennrich2016NeuralMT} to split words into sub-words. The size of sub-words is set to 30K for each language.

For the WMT corpus, we evaluate our approach on a Spanish-French (Es-Fr) translation task with a zero-resource setting. We combine the following corpora to form the Es-En and En-Fr parallel corpora: Common Crawl, News Commentary, Europarl v7 and UN. All the sentences are tokenized by the \verb|tokenize.perl| script. Newstest2011 serves as the development set and Newstest2012 and Newstest2013 serve as test sets. We use case-sensitive BLEU to evaluate translation results. BPE is also used to reduce the vocabulary size. The size of sub-words is set to 43K, 33K, 43K for Spanish, English and French, respectively. See Table \ref{table:D1} for detailed statistics for the Europarl and WMT corpora.


\begin{table*}[!t]
	\centering
	\begin{tabular}{l | l | c | c | c | c | c }
	& \multirow{2}{*}{Approx.}  & \multicolumn{5}{c}{Iterations}  \\
	\cline{3-7}
	&  & 0 & 2w & 4w & 6w & 8w  \\
	\hline \hline
	\multirow{2}{*}{$\mathcal{J}_{\mathrm{SENT}}$} & greedy & 313.0 & 73.1 & 61.5 & 56.8 & 55.1  \\
	& beam & 323.5 & 73.1 & 60.7 & 55.4 & 54.0 \\	
	\hline
	\multirow{3}{*}{$\mathcal{J}_{\mathrm{WORD}}$} & greedy & 274.0 & 51.5 & 43.1 & 39.4 & 38.8  \\
	& beam & 288.7 & 52.7 & 43.3 & 39.2 & 38.4  \\
	& sampling  &  268.6 & 53.8 & 46.6 & 42.8 & 42.4  \\
	\end{tabular}
	\caption{Verification of sentence-level and word-level assumptions by evaluating approximated KL divergence from the source-to-target model to the pivot-to-target model over training iterations of the source-to-target model. The pivot-to-target model is trained and kept fixed. } \label{table:VF}
\end{table*}

\begin{table*}[!t]
	\centering
	\begin{tabular}{c | l | c | c  }
		& Method & \hfil Es$\rightarrow$ Fr & \hfil De$\rightarrow$ Fr  \\
		\hline \hline
		\multirow{4}{*}{Cheng et al. \shortcite{Cheng2016NeuralMT}} & pivot & \hfil 29.79  & \hfil 23.70   \\
		& hard  & \hfil 29.93  & \hfil 23.88   \\	
		& soft &  \hfil 30.57  & \hfil 23.79 \\
		& likelihood  &  \hfil 32.59 &  \hfil 25.93 \\
		\hline
		\multirow{2}{*}{Ours} & sent-beam  &  \hfil 31.64 & \hfil 24.39 \\
		& word-sampling  & \hfil 33.86 & \hfil 27.03 \\
	\end{tabular}
	\caption{Comparison with previous work on Spanish-French and German-French translation tasks from the Europarl corpus. English is treated as the pivot language. The likelihood method uses 100K parallel source-target sentences, which are not available for other methods.} \label{table:E1}
\end{table*}

We leverage an open-source NMT toolkit {\em dl4mt} implemented by Theano \footnote{\textit{dl4mt-tutorial}: https://github.com/nyu-dl} for all the experiments  and compare our approach with state-of-the-art multilingual methods \cite{Firat2016ZeroResourceTW} and pivot-based methods \cite{Cheng2016NeuralMT}. Two variations of our framework are used in the experiments:
\begin{enumerate}
	\item Sentence-Level Teaching: for simplicity, we use the mode as suggested in \cite{Kim2016SequenceLevelKD} to approximate the target sentence space in calculating the expected gradients with respect to the expectation in Equation (\ref{sentence:final}). We run beam search on the pivot sentence with the teacher model and choose the highest-scoring target sentence as the mode. Beam size with $k=1$ (greedy decoding) and $k=5$ are investigated in our experiments, denoted as \textit{sent-greedy} and \textit{sent-beam}, respectively.\footnote{We can also adopt sampling and $k$-best list for approximation. Random sampling brings a large variance \cite{Sutskever2014SequenceTS, Ranzato2015SequenceLT, He2016DualLF} for sentence-level teaching. For $k$-best list, we renormalize the probabilities
	$$P(\mathbf{y}|\mathbf{z}; \hat{\bm{\theta}}_{z \rightarrow y}) \sim \frac{P(\mathbf{y}|\mathbf{z}; \hat{\bm{\theta}}_{z \rightarrow y})^\alpha}{\sum_{\mathbf{y}\in \mathcal{Y}_k}P(\mathbf{y}|\mathbf{z}; \hat{\bm{\theta}}_{z \rightarrow y})^\alpha },$$ where $\mathcal{Y}_k$ is the $k$-best list from beam search of the teacher model and $\alpha$ is a hyperparameter controling the sharpness of the distribution \cite{Och2003MinimumER}. We set $k=5$ and $\alpha=5\times10^{-3}$. The results on test set for Eureparl Corpus are 32.24 BLEU over Spanish-French translation and 24.91 BLEU over German-French translation, which are slightly better than the sent-beam method. However, considering the traing time and the memory consumption, we believe mode approximation is already a good way to approximate the target sentence space for sentence-level teaching.}
	\item Word-Level Teaching: we use the same mode approximation approach as in sentence-level teaching to approximate the expectation in Equation \ref{word:final}, denoted as \textit{word-greedy} (beam search with $k=1$) and \textit{word-beam} (beam search with $k=5$), respectively. Besides, Monte Carlo estimation by sampling from the teacher model is also investigated since it introduces more diverse data, denoted as \textit{word-sampling}. 
\end{enumerate}

\subsection{Assumptions Verification}\label{4.2}
To verify the assumptions in Section \ref{C1}, we train a source-to-target translation model $P({\mathbf{y}|\mathbf{x};\bm{\theta}_{x\rightarrow y}})$ and a pivot-to-target translation model $P({\mathbf{y}|\mathbf{z};\bm{\theta}_{z\rightarrow y}})$ using the trilingual Europarl corpus. Then, we measure the sentence-level and word-level KL divergence from the source-to-target model $P({\mathbf{y}|\mathbf{x};\bm{\theta}_{x\rightarrow y}})$ at different iterations to the trained pivot-to-target model $P({\mathbf{y}|\mathbf{z};\hat {\bm{\theta}}_{z\rightarrow y}})$ by caculating $\mathcal{J}_{\mathrm{SENT}}$ (Equation (\ref{eq:sent})) and $\mathcal{J}_{\mathrm{WORD}}$ (Equation (\ref{eq:word})) on 2,000 parallel source-pivot sentences from the development set of WMT 2006 shared task. 

Table \ref{table:VF} shows the results. The source-to-target model is randomly initialized at iteration 0. We find that $\mathcal{J}_{\mathrm{SENT}}$ and $\mathcal{J}_{\mathrm{WORD}}$ decrease over time, suggesting that the source-to-target and pivot-to-target models do have small KL divergence at both sentence and word levels. 

\subsection{Results on the Europarl Corpus}\label{4.3}

Table \ref{table:E1} gives BLEU scores on the Europarl corpus of our best performing sentence-level method (sent-beam) and word-level method (word-sampling) compared with pivot-based methods \cite{Cheng2016NeuralMT}. We use the same data preprocessing as in \cite{Cheng2016NeuralMT}. We find that both the sent-beam and word-sampling methods outperform the pivot-based approaches in a zero-resource scenario across language pairs. Our word-sampling method improves over the best performing zero-resource pivot-based method (soft) on Spanish-French translation by +3.29 BLEU points and German-French translation by +3.24 BLEU points. In addition, the word-sampling mothod surprisingly obtains improvement over the likelihood method, which leverages a source-target parallel corpus. The significant improvements can be explained by the error propagation problem of pivot-based methods, which propagates translation error of the source-to-pivot translation process to the pivot-to-target translation process. 
\begin{figure*}[!t]
	\centering\includegraphics[width=6.3in]{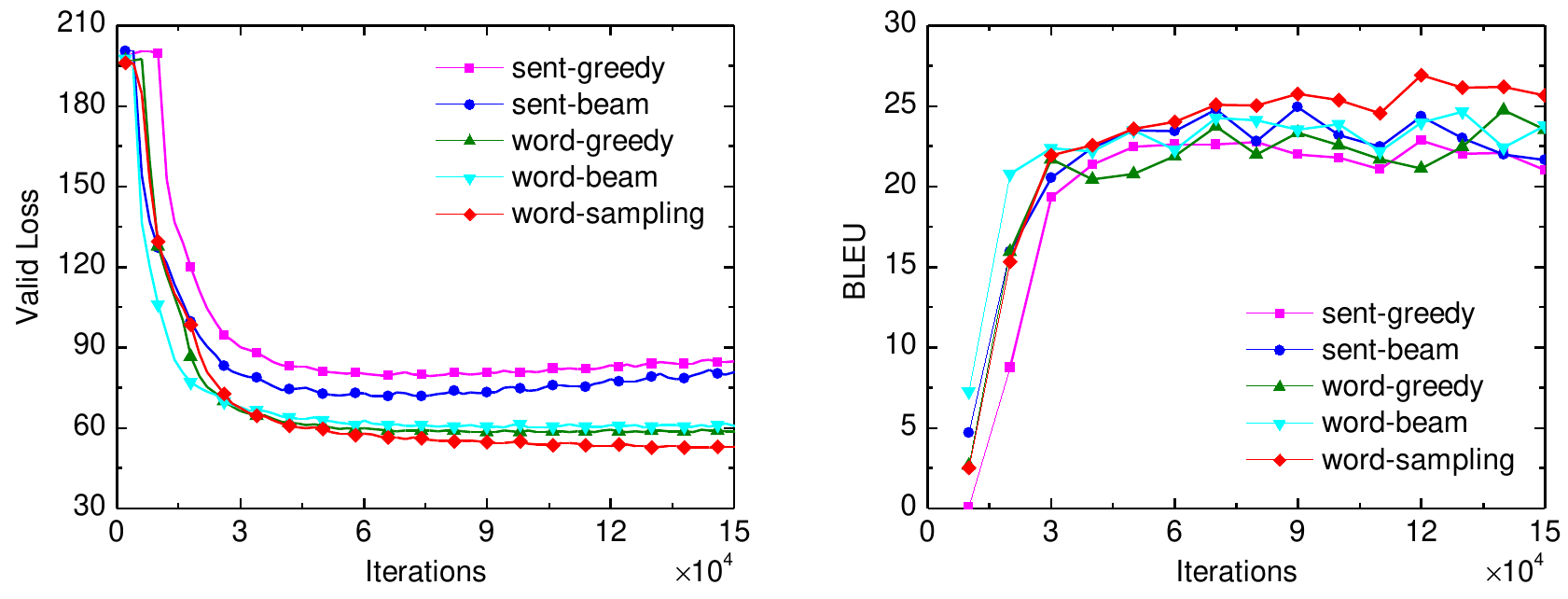}
	\vspace{-10pt}
	\caption{Validation loss and BLEU across iterations of our proposed methods.}\label{fig:E1}
\end{figure*}
\begin{table}[!t]
	\centering
	\begin{tabular}{l | c | c | c | c }
		\multirow{2}{*}{Method} &  \multicolumn{2}{|c|}{Es$\rightarrow$ Fr}  & \multicolumn{2}{c}{De$\rightarrow$ Fr}  \\
		\cline{2-5}
		& dev  & test & dev & test  \\
		\hline \hline
		sent-greedy & 31.00  & 31.05 & 22.34 & 21.88 \\
		sent-beam & 31.57  & 31.64 & 24.95 & 24.39 \\
		\hline
		word-greedy & 31.37  & 31.92 & 24.72 & 25.15 \\
		word-beam & 30.81  & 31.21 & 24.64 & 24.19 \\
		word-sampling & 33.65  & 33.86 & 26.99 & 27.03 \\
	\end{tabular}
	\caption{Comparison of our proposed methods on Spanish-French and German-French translation tasks from the Europarl corpus. English is treated as the pivot language.} \vspace{-10pt}\label{table:E2}
\end{table}
\begin{table*}[!t]
	\centering
	\begin{tabular}{l | l | r | r | c | c  | c }
		& \multirow{2}{*}{Method} & \multicolumn{3}{|c|}{Training}  & \multicolumn{2}{c}{BLEU} \\
		& & Es$\rightarrow$ En & En$\rightarrow$ Fr & Es$\rightarrow$ Fr & \small{Newstest2012} & \small{Newstest2013} \\
		\hline \hline
		\multicolumn{7}{c}{{\em Existing zero-resource NMT systems}} \\
		\hline
		Cheng et al. \shortcite{Cheng2016NeuralMT}$^\dagger$ & pivot & 6.78M & 9.29M & - & 24.60 & - \\
		Cheng et al. \shortcite{Cheng2016NeuralMT}$^\dagger$ & likelihood & 6.78M & 9.29M & 100K & 25.78 & - \\
		Firat et al. \shortcite{Firat2016ZeroResourceTW} & one-to-one & 34.71M & 65.77M & - & 17.59 & 17.61 \\
		Firat et al. \shortcite{Firat2016ZeroResourceTW}$^\dagger$ & many-to-one & 34.71M & 65.77M & - & 21.33 & 21.19 \\
		\hline
		\multicolumn{7}{c}{{\em Our zero-resource NMT system}} \\
		\hline
		& word-sampling & 6.78M & 9.29M & - & 28.06 & 27.03
		
	\end{tabular}
	\caption{Comparison with previous work on Spanish-French translation in a zero-resource scenario over the WMT corpus. The BLEU scores are case sensitive. $\dagger$: the method depends on two-step decoding.} \vspace{-10pt} \label{table:E3}
\end{table*}
%

Table \ref{table:E2} shows BLEU scores on the Europarl corpus of our five proposed methods. For sentence-level approaches, the sent-beam method outperforms the sent-greedy method by +0.59 BLEU points over Spanish-French translation and +2.51 BLEU points over German-French translation on the test set. The results are in line with our observation in Table \ref{table:VF} that sentence-level KL divergence by beam approximation is smaller than that by greedy approximation. However, as the time complexity grows linearly with the number of beams $k$, the better performance is achieved at the expense of search time. 

For word-level experiments, we observe that the word-sampling method performs much better than the other two methods: +1.94 BLEU points on Spanish-French translation and +1.88 BLEU points on German-French translation over the word-greedy method; +2.65 BLEU points on Spanish-French translation and +2.84 BLEU points on German-French translation over the word-beam method. Although Table \ref{table:VF} shows that word-level KL divergence approximated by sampling is larger than that by greedy or beam, sampling approximation introduces more data diversity for training, which dominates the effect of KL divergence difference. 
\begin{table*}[!t]
	\small
	\centering
	\begin{tabular}{l|l|p{1.5\columnwidth}}
		\hline
		\multirow{6}{*}{groundtruth}  &\multirow{2}{*}{source} & Os sent\'ais al volante en la costa oeste , en San Francisco , y vuestra misi\'on es llegar los primeros a Nueva York . \\
		\cline{2-3} &\multirow{2}{*}{pivot} & You get in the car on the west coast , in San Francisco , and your task is to be the first one to reach New York .\\
		\cline{2-3} &\multirow{2}{*}{target} & Vous vous asseyez derri\`ere le volant sur la c\^ote ouest \`a San Francisco et votre mission est d\&apos; arriver le premier \`a New York . \\
		\hline
		\multirow{4}{*}{pivot}  &\multirow{2}{*}{pivot} & You \&apos;ll feel at \textcolor{blue}{\textit{the west coast}} \textcolor{blue}{\textit{in San Francisco , and your}} mission \textcolor{blue}{\textit{is to}} get \textcolor{blue}{\textit{the first}} to \textcolor{blue}{\textit{New York .}} [BLEU: 33.93] \\
		\cline{2-3} &\multirow{2}{*}{target} & \textcolor{blue}{\textit{Vous vous}} sentirez comme chez vous \textcolor{blue}{\textit{\`a San Francisco}} , \textcolor{blue}{\textit{et votre mission est d\&apos;}} obtenir \textcolor{blue}{\textit{le premier \`a New York .}} [BLEU: 44.52]\\
		\hline
		\multirow{4}{*}{likelihood}  &\multirow{2}{*}{pivot} & You feel at \textcolor{blue}{\textit{the west coast , in San Francisco , and your}} mission \textcolor{blue}{\textit{is to}} reach \textcolor{blue}{\textit{the first}} to \textcolor{blue}{\textit{New York .}} [BLEU: 47.22]\\
		\cline{2-3} &\multirow{2}{*}{target} & \textcolor{blue}{\textit{Vous vous}} sentez \`a \textcolor{blue}{\textit{la c\^ote ouest}} , \textcolor{blue}{\textit{\`a San Francisco}} , \textcolor{blue}{\textit{et votre mission est d\&apos;}} atteindre \textcolor{blue}{\textit{le premier \`a New York .}} [BLEU: 49.44]\\
		\hline
		\multirow{2}{*}{word-sampling}  &\multirow{2}{*}{target} & \textcolor{blue}{\textit{Vous vous}} sentez au \textcolor{blue}{\textit{volant sur la c\^ote ouest}} , \textcolor{blue}{\textit{\`a San Francisco et votre mission est d\&apos; arriver le premier \`a New York .}} [BLEU: 78.78]\\
		\hline
	\end{tabular}
	\caption{Examples and corresponding sentence BLEU scores of translations using the pivot and likelihood methods in \cite{Cheng2016NeuralMT} and the proposed word-sampling method.  We observe that our approach generates better translations than the methods in \cite{Cheng2016NeuralMT}. We italicize \protect \textcolor{blue}{\textit{correct translation segments}} which are no short than 2-grams.} \label{table:example}
\end{table*}

We plot validation loss\footnote{\textit{Validation loss}: the average negative log-likelihood of sentence pairs on the validation set.} and BLEU scores over iterations on the German-French translation task in Figure \ref{fig:E1}. We observe that word-level models tend to have lower validation loss compared with sentence-level methods. Generally, models with lower validation loss tend to have higher BLEU. Our results indicate that this is not necessarily the case: the sent-beam method converges to +0.31 BLEU points on the validation set with +13 validation loss compared with the word-beam method. Kim and Rush \shortcite{Kim2016SequenceLevelKD} claim a similar observation in data distillation for NMT and provide an explanation that student distributions are more peaked for sentence-level methods. This is indeed the case in our result: on German-French translation task the argmax for the sent-beam student model (on average) approximately accounts for 3.49\% of the total probability mass, while the corresponding number is 1.25\% for the word-beam student model and 2.60\% for the teacher model.

\subsection{Results on the WMT Corpus}

The word-sampling method obtains the best performance in our five proposed approaches according to experiments on the Europarl corpus. To further verify this approach, we conduct experiments on the large scale WMT corpus for Spanish-French translation. Table \ref{table:E3} shows the results of our word-sampling method in comparison with other state-of-the-art baselines. Cheng et al. \shortcite{Cheng2016NeuralMT} use the same datasets and the same preprocessing as ours. Firat et al. \shortcite{Firat2016ZeroResourceTW} utilize a much larger training set.\footnote{Their training set does not include the Common Crawl corpus.} Our method obtains significant improvement over the pivot baseline by +3.46 BLEU points on Newstest2012 and over many-to-one by +5.84 BLEU points on Newstest2013. Note that both methods depend on a source-pivot-target decoding path. Table \ref{table:example} shows translation examples of the pivot and likelihood methods proposed in \cite{Cheng2016NeuralMT} and our proposed word-sampling method. For the pivot and likelihood methods, the Spainish sentence segment 'sent\'ais al volante' is lost when translated to English. Therefore, both methods miss this information in the translated French sentence. However, the word-sampling method generates 'volant sur', which partially translates 'sent\'ais al volante', resulting in improved translation quality of the target-language sentence.

\begin{table}[!t]
	\centering
	\begin{tabular}{ l | c | c | c | c }
		\multirow{2}{*}{Method} & \multicolumn{3}{c|}{Corpus} &  \multirow{2}{*}{BLEU} \\
		\cline{2-4}
		& De-En & De-Fr & En-Fr &  \\
		\hline 
		MLE & $\times$  & $\surd$ & $\times$ & 19.30 \\
		transfer & $\times $  & $\surd$ & $\surd$ & 22.39 \\ %
		pivot & $\surd$  & $\times$ & $\surd$ & 17.32 \\
		\hline
		Ours & $\surd$  & $\times$ & $\surd$ & 22.95 \\
	\end{tabular}
	\caption{Comparison on German-French translation task from the Europarl corpus with 100K German-English sentences. English is regarded as the pivot language. Transfer represents the transfer learning method in \cite{Zoph2016TransferLF}. 100K parallel German-French sentences are used for the MLE and transfer methods.}\vspace{-10pt}\label{table:E4}
\end{table}
\subsection{Results with Small Source-Pivot Data}
The word-sampling method can also be applied to zero-resource NMT with a small source-pivot corpus. Specifically, the size of the source-pivot corpus is orders of magnitude smaller than that of the pivot-target corpus. This setting makes sense in applications. For example, there are significantly fewer Urdu-English corpora available than English-French corpora.

To fulfill this task, we combine our best performing word-sampling method with the initialization and parameter freezing strategy proposed in \cite{Zoph2016TransferLF}. The Europarl corpus is used in the experiments. We set the size of German-English training data to 100K and use the same teacher model trained with 900K English-French sentences. 

Table \ref{table:E4} gives the BLEU score of our method on German-French translation compared with three other methods. Note that our task is much harder than transfer learning \cite{Zoph2016TransferLF} since the latter depends on a parallel German-French corpus. Surprisingly, our method outperforms all other methods. We significantly improve the baseline pivot method by +5.63 BLEU points and the state-of-the-art transfer learning method by +0.56 BLEU points. 

\section{Related Work}
Training NMT models in a zero-resource scenario by leveraging other languages has attracted intensive attention in recent years. Firat et al. \shortcite{Firat2016ZeroResourceTW} proposed an approach which delivers the multi-way, multilingual NMT model proposed by \cite{Firat2016MultiWayMN} for zero-resource translation. They used the multi-way NMT model trained by other language pairs to generate a pseudo parallel corpus and fine-tuned the attention mechanism of the multi-way NMT model to enable zero-resource translation. Several authors proposed a universal encoder-decoder network in multilingual scenarios to perform zero-shot learning \cite{Johnson2016GooglesMN,Ha2016TowardMN}. This universal model extracts translation knowledge from multiple different languages, making zero-resource translation feasible without direct training. 

Besides multilingual NMT, another important line of research attempts to bridge source and target languages via a pivot language. This idea is widely used in SMT \cite{de2006catalan, Cohn2007MachineTB, Utiyama2007ACO, Wu2007PivotLA, Bertoldi2008PhrasebasedSM, Wu2009RevisitingPL, Zahabi2013UsingCV, Kholy2013LanguageIC}. Cheng et al. \shortcite{Cheng2016NeuralMT} propose pivot-based NMT by simultaneously improving source-to-pivot and pivot-to-target translation quality in order to improve source-to-target translation quality. Nakayama and Nishida \shortcite{Nakayama2016ZeroresourceMT} achieve zero-resource machine translation by utilizing image as a pivot and training multimodal encoders to share common semantic representation.

Our work is also related to knowledge distillation, which trains a compact model to approximate the function learned by a larger, more complex model or an ensemble of models \cite{Bucila2006ModelC,Ba2014DoDN,Li2014LearningSD,Hinton2015DistillingTK}.
Kim and Rush \shortcite{Kim2016SequenceLevelKD} first introduce knowledge distillation in neural machine translation. They suggest to generate a pseudo corpus to train the student network. Compared with their work, we focus on zero-resource learning instead of model compression.

\section{Conclusion}
In this paper, we propose a novel framework to train the student model without parallel corpora available under the guidance of the pre-trained teacher model on a source-pivot parallel corpus. We introduce sentence-level and word-level teaching to guide the learning process of the student model. Experiments on the Europarl and WMT corpora across languages show that our proposed word-level sampling method can significantly outperforms the state-of-the-art pivot-based methods and multilingual methods in terms of translation quality and decoding efficiency.

We also analyze zero-resource translation with small source-pivot data, and combine our word-level sampling method with initialization and parameter freezing suggested by \cite{Zoph2016TransferLF}. The experiments on the Europarl corpus show that our approach obtains an significant improvement over the pivot-based baseline.

In the future, we plan to test our approach on more diverse language pairs, e.g., zero-resource Uyghur-English translation using Chinese as a pivot. It is also interesting to extend the teacher-student framework to other cross-lingual NLP applications as our method is transparent to architectures.


\section*{Acknowledgments}
This work was done while Yun Chen is visiting Tsinghua University. This work is partially supported by the National Natural Science Foundation of China (No.61522204, No. 61331013) and the 863 Program (2015AA015407).

\bibliography{acl2017}
\bibliographystyle{acl_natbib}
\end{document}